\documentclass{article}

\usepackage{xcolor}

\usepackage{microtype}
\usepackage{graphicx}
\usepackage{subfigure}
\usepackage{booktabs} 
\usepackage{etoolbox}

\usepackage{hyperref}

\usepackage[accepted]{custom2019}

\customtitlerunning{Handling many conversions per click in modeling delayed feedback.}

\begin{document}

\twocolumn[
\customtitle{Handling many conversions per click in modeling delayed feedback}

\begin{customauthorlist}
\customauthor{Ashwinkumar Badanidiyuru}{googleResearch}
\customauthor{Andrew Evdokimov}{google}
\customauthor{Vinodh Krishnan}{google}
\customauthor{Pan Li}{google}
\customauthor{Wynn Vonnegut}{google}
\customauthor{Jayden Wang}{google}
\\
$^1$Google Research, Mountain View, USA. 
\\
$^2$Google, Mountain View, USA
\end{customauthorlist}

\customcorrespondingauthor{Ashwinkumar Badanidiyuru}{ashwinkumarbv@google.com}

\vskip 0.3in
]

\printAffiliationsAndNotice

\begin{abstract}
Predicting the expected value or number of post-click conversions (purchases or other events) is a key task in performance-based digital advertising. In training a conversion optimizer model, one of the most crucial aspects is handling delayed feedback with respect to conversions, which can happen multiple times with varying delay. This task is difficult, as the delay distribution is different for each advertiser, is long-tailed, often does not follow any particular class of parametric distributions, and can change over time. We tackle these challenges using an unbiased estimation model based on three core ideas. The first idea is to split the label as a sum of labels with different delay buckets, each of which trains only on mature label, the second is to use thermometer encoding to increase accuracy and reduce inference cost, and the third is to use auxiliary information to increase the stability of the model and to handle drift in the distribution.
\end{abstract}

\section{Introduction}
For decades, the advertising industry has used reach or number of users who viewed an advertisement, to estimate the effectiveness of ads and to price advertising opportunities. This covered all forms of advertising including advertising on billboards, newspapers and television. Online advertising also started out following this model, charging advertisers for the number of times their ad was viewed, and even now such arrangements remain popular, especially for brand advertising.

However, increasingly, the growing sophistication of online advertising platforms allows advertisers to automatically target deeper goals as well. In the 1990s, we saw the emergence of cost-per-click advertising and finer-grained targeting, so advertisers would only be charged if a user clicked on their ad, and were able to manually assign distinct values to different segments of users. Combined with real-time bidding (RTB) for deciding which ad to show, this necessitated the creation of click-through rate prediction models to compute the probability of a click on a particular impression for a particular advertiser.

In recent years, we have seen two new, synergistic emerging trends. First, increasing automation in targeting means that while advertisers would previously need to specify the exact keywords or sets of users they wish to see their ads, now advertising platforms are increasingly able to automatically find impressions that would contribute to satisfying the advertisers' objectives, with significantly reduced manual targeting configuration. At the same time, platforms are increasingly allowing advertisers to provide more details about their objectives by specifying what events (for example, purchases) they value after a user has clicked.

One simple advertising campaign setup for achieving this is for the advertiser to specify the events they care about, and for the campaign objective to be to maximize the number of events that are done by users who click on advertisements, under the constraints of maximum total spend and average cost per event. A more sophisticated form of this configuration is to additionally specify a value for each event that occurs, and for the objective to be to maximize the overall value delivered by the advertising campaign. On the advertising platform side, optimizing for these objectives then requires the prediction of number or value of post-click events (commonly called ``conversions"), in addition to the prediction of the click-through rate as before.

Advertisers can set up once per click (OPC) or many per click (MPC) types of conversions. In the OPC case, at most one conversion following the click will be counted, while in the MPC case, a click can have any number of conversions as long as they occur within the post-click window. The duration of the post-click attribution window is set by the advertiser and can be as low as 2 hours or as long as 90 days. 

For example, a ride sharing app may have a campaign that optimizes only for users who actually go on to take a first ride (OPC). On the other hand, a puzzle game could report a conversion whenever a user reaches level 20 in the game (OPC) or makes an in-app purchase (MPC).

Building a conversion optimizer model to estimate the expected number of conversions for MPC or OPC/MPC mixed campaigns raises several challenges that are absent in estimating the click-through rate of pure OPC conversions.\\
1. The events that are being predicted can take up to 90 days to be reported.\\
2. Unlike clicks or OPC conversions, there can be (and often are) multiple events associated with each click or impression. This also increases the difficulty of the previous challenge, since we cannot be sure that we have seen all events attributable to the click until the end of the attribution window.\\
3. Since events can be defined by advertisers in arbitrary ways, their distributions are highly heterogeneous, both in terms of the rate and delay of events.\\
4.The environment is non-stationary, with user behavior changing in response to changes on advertiser apps or websites, advertisers sometimes changing their optimization objectives, and new advertising campaigns with new objectives being created. In particular, for new campaigns, very little generalization is possible, since we do not know \emph{a priori} how the events will be defined.

To address this last concern of non-stationarity, it is common practice to use online training for models, training them on data that is as close to real-time as possible. In this context, however, the problem of conversion delay becomes especially acute: when training close to real time, the model misses any conversions that would arrive after its training delay, and so sees systematically fewer conversions than will ultimately happen. Although the conversion delay problem is also a challenge for batch training (as some portion of examples will still be immature), we focus on the online training setting in this paper. Our results trivially generalize to batch training.

In this work, we develop a model for the expected number of conversions that is able to train close to real time, while achieving neutral long-term bias of predictions, largely eliminating the trade-off between the accuracy gained through short training delay and the bias in longer-delay examples. To do so, we model the expected distribution of conversion delays in a non-parametric way, while taking advantage of any available intermediate information to improve the accuracy of predictions.
\section{Related work on delay modeling}
Delay modeling is closely related to the problem of censored feedback. There has been an extensive literature on this topic and one can find related work in \cite{EL2008} and it's references. While this literature is quite relevant it's not directly applicable. On the other hand there have also been multiple papers such as \cite{Agarwal2010,MenonCGAK11,LeeODL12,Rosales2012} studying conversion optimization, but these do not address the problem of delayed feedback in labels.  

The first formal study of handling delayed feedback in conversion optimizer was initiated by \cite{Chapelle2014A2}. The results in this paper, as well as all following papers on this problem have been restricted to the special case of at most one conversion per click. The solution proposed in \cite{Chapelle2014A2} uses many of the tools in survival analysis. At a high level the authors train a model to estimate the conversion delay (time from click to conversion) assuming that it follows an exponential distribution. They then compute the probability of observing a conversion given the click age as a function of the probability of conversion and the delay distribution using Bayes' rule. Finally, they train a model for the probability of observing a conversion given the click age on the realized label.

There have been several follow up papers to \cite{Chapelle2014A2} in the same restricted setting. \cite{Hubbard2019} improves the result by assuming that the delay distribution is a geometric distribution with a beta prior, \cite{Ktena2019} studies the problem with different types of loss functions such as inverse propensity scoring and importance sampling, \cite{Yoshikawa2018} considers a non-parametric family of delay distributions which is a weighted sum of kernel values, \cite{Mann2019} studies it in the adversarial online-learning framework, \cite{Wendi2017} studies it with the Weibull family of distributions, \cite{Safari2017} gives a biased estimator for improving efficiency, and \cite{Vernade2017} studies it in the bandit setting. These solutions aren't able to handle the more general case of multiple conversions per click. 

There are recent works handling this problem without the assumption of a parametric delay distribution. \cite{Saito2020} proposed a dual learning algorithm of the CVR predictor and bias estimator. \cite{Yasui2020} addressed this problem by using an importance weighting approach typically used for covariate shift correction. \cite{Su2020} calibrated the delay model by learning a dynamic hazard function with the post-click data, and \cite{Kato2020} proposed a method with an unbiased and convex empirical risk constructed from samples. These works also do not extend to the MPC case.

The first paper to study handling multiple conversions per click was \cite{Choi2020}. They extend \cite{Chapelle2014A2} by assuming that conversions come as a negative binomial distribution with exponentially distributed time delays between them. There are two challenges with this approach. The first is that it works only for integer conversions and cannot easily extend to predicting the expected value where the label is float. The second, as noted in the paper, is the loss function they define is non-convex and the model can either predict that the data has high conversion rate and long delay or low conversion rate and short delay. Empirically, this works out fine in batch training as the model sees some mature data in each batch to resolve this issue. Unfortunately, in online learning this is not the case, as the model stops seeing mature data when training on more recent examples. This makes non-parametric methods like the one proposed in this paper more robust in these settings. 

\section{Preliminaries}
We can more formally describe the problem as follows:

Our model trains on examples in the time range $[0, t)$ in sequence, visiting each example once. Each example $p$ is associated with two types of features. Features $X_p$ are available at the time the example would have been predicted on in serving, while features $L_p$ are delayed and may change from time $t_p$ (the time at which we would have been required to make a prediction on this example) to time $t_p + M$ for a fixed $M$, after which we disregard any further updates to the example. The label, $y_p \in [0, \infty)$ can be considered a sub-type of features $L_p$. For convenience, we will say that the label $y_p$ is the total number or value of individual events $\{y_{p,0}, y_{p,1}, ...\}$ that become visible at times $\{t_{p,0}, t_{p,1}, ...\}$, although our approach could also be used for fully continuous response variables.

Finally, at time $t$, we see an example with features $X_p$, and must make a prediction $\bar{y} = f(X_p, \theta)$ for it using model parameters $\theta$. We desire this prediction to be accurate and unbiased under the assumption that the time-distribution of the label and features $L_p$ conditional on the input features is relatively stationary.

\section{Model}
\subsection{Core ideas}

The fundamental goal of our approach is to model the expected delay distribution for each particular example. From this sub-model, we want to receive a prediction, such that when we see an incomplete label (i.e. one which may still be updated upwards over time), we can predict what the label will be when it is complete, and then use this completed label for training the main model.

To model the delay distribution, we introduce a novel model configuration, based on three core ideas.

\paragraph*{Split the label into different delay buckets}
Say we have a training example that came from time $t_p$ and want to train on it at time $t_p + K$, $K < M$. Then, we know the portion of the label that occurred in $[t_p, t_p + K)$, and in order to avoid bias in training, need to predict the ``remaining" portion of the label for the time period $[t_p + K, t_p + M)$.

To make this prediction, there are two reasonable approaches that could be taken: either we can attempt to make a model that can make this prediction for arbitrary values of $K$ or choose a number of fixed values at which we're able to make predictions. The first approach requires either a parametric distribution assumption, which is undesirable due the heterogeneity of our delay distribution, or the use of a detailed time series model. So, we elect to use the second approach to avoid unnecessary complexity.

To do so, we interpret the overall label as a sum of conversions that fall in different delay buckets $[t_p, t_p + d_1), [t_p + d_1, t_p + d_2), \ldots, [t_p + d_n, t_p + M)$. Then we can create a set of sub-models $f_i(X_p)$ to predict the label in each bucket $[t_p + d_i,t_p + d_{i+1})$, where $d_0 = 0$, and train each model only on examples whose age is at least $t_p + d_{i+1}$ (i.e. examples for which the label of the bucket is complete).

Then, if at time $t_k$ we wish to estimate the complete label for example $p$, we can find the latest sub-model $f_m$ the beginning of whose prediction interval precedes time $t_k$, and compute the total expected label as the sum of the truncated known label part and the predicted label part: $y'_{p,t_k} = |{y_{p,i} : t_{p,i} < t_p + d_m}| + \sum_{i \ge m} f_i(X_p)$. It's easy to see that $y'_{p,.}$ is an unbiased estimator for $y$ if and only if each sub-model $f_i$ is an unbiased estimator for the label part on its own interval.

\paragraph*{Thermometer encoding of the labels}

Observe that in the above setup, any observed events ${y_{p,j} : t_p + d_m < j < t_k}$ are discarded when computing the predicted label, because they overlap with the time bucket of a sub-model that we need to use. Because of this, it's important to have a sufficiently large number of sub-models that the proportion of such discarded events is sufficiently small. However, as we increase the number of models, each one is responsible for shorter and short intervals, and so has fewer and fewer positive labels.

To address this label sparsity issue, we introduce a thermometer encoding of the label: instead of separating the label into non-overlapping buckets as above, we separate it into overlapping buckets $[t_p, t_p + M), [t_p + d_1, t_p + M), ..., [t_p + d_n, t_p + M)$, so that the label for each sub-model is the number of events occurring from its beginning time to the end of the full time period. This way, almost regardless of how many sub-models we use, each will still cover a reasonable time range, and avoid label sparsity.

Using this label encoding also reduces the cost of inference. If each sub-model is responsible for a narrow time bucket, we must evaluate the sub-models covering all following time periods to complete the label or evaluate all sub-models to get the full time range prediction. By using thermometer encoding, we need to evaluate only a single one.

\paragraph*{Auxiliary information}

The third key idea is to use the delayed features $L_p$ as sub-model inputs to improve the accuracy and bias of the label completion. In particular, we provide as an input the ``label so far", i.e. a feature describing the labels in $[t_p, t_p + d_i)$, to each sub-model $f_i$. We can draw an analogy with conditional entropy to note that $H(Y|X) = H(Y) - I(X, Y)$, so conditioning our prediction on the label so far will reduce its entropy when they have high mutual information, which is shown in Figure 1. 

\begin{figure}
    \centering
    \includegraphics[width=8cm]{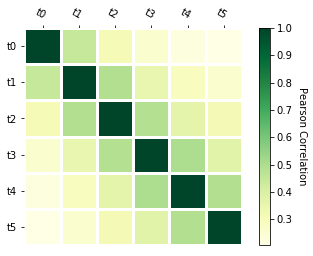}
    \caption{The correlation of number of conversions between different delay buckets. The high correlation between buckets indicates the auxiliary information will help sub-model prediction}
\end{figure}

In particular, note that providing this auxiliary information helps support the assumption that the delay distribution should be stationary in time conditional on the model inputs. For example, if user behaviors where the first event comes later become more frequent over time, the sub-models could correctly adjust for this drift when performing label completion even if the change is not associated with the standard input features.

Observe also that any auxiliary information we use in training must also be available when performing inference on the sub-model. This requirement works well with thermometer label encoding: we can provide sub-model $f_i$ any information available up to $t_p + d_i$, since we only use it for inference after that point in time. If, however, we wanted to provide auxiliary information to models using bucket encoding, we could only provide information available up to $t_p + d_1$, since we may use every model to complete an ``early" partial label.

\subsection{Setup details}

We give exact training setup succinctly with a list of models which are trained, their features/labels and examples that they train on in Table \ref{table:ModelsFeaturesFilters}. This model setup requires careful handling of several details.

First, in the training regime described, where examples are visited once sequentially, observe that for an example at time $t_p + k$, $k < d_i$, only sub-model predictions $f_j : j < i$ will be trained. As a result, if all predictions are coming from a single model, predictions which are not being trained may be affected by those that are, and the model may experience catastrophic forgetting. To mitigate this effect in this training regime, it is desirable to have fully separate sub-models for each of the predictions.

This then adds an extra constraint when deciding the number of buckets for the labels, and so the number of predictions. If we choose a low number of buckets, we may lose information, since when training, we discard label parts that occur part-way through a bucket (and instead substitute a prediction for them). If we choose a larger number of buckets, the training cost of the overall model increases. In practice, we have found that 3-10 buckets can be a reasonable choice.

\begin{figure}[h]
\includegraphics[width=8cm]{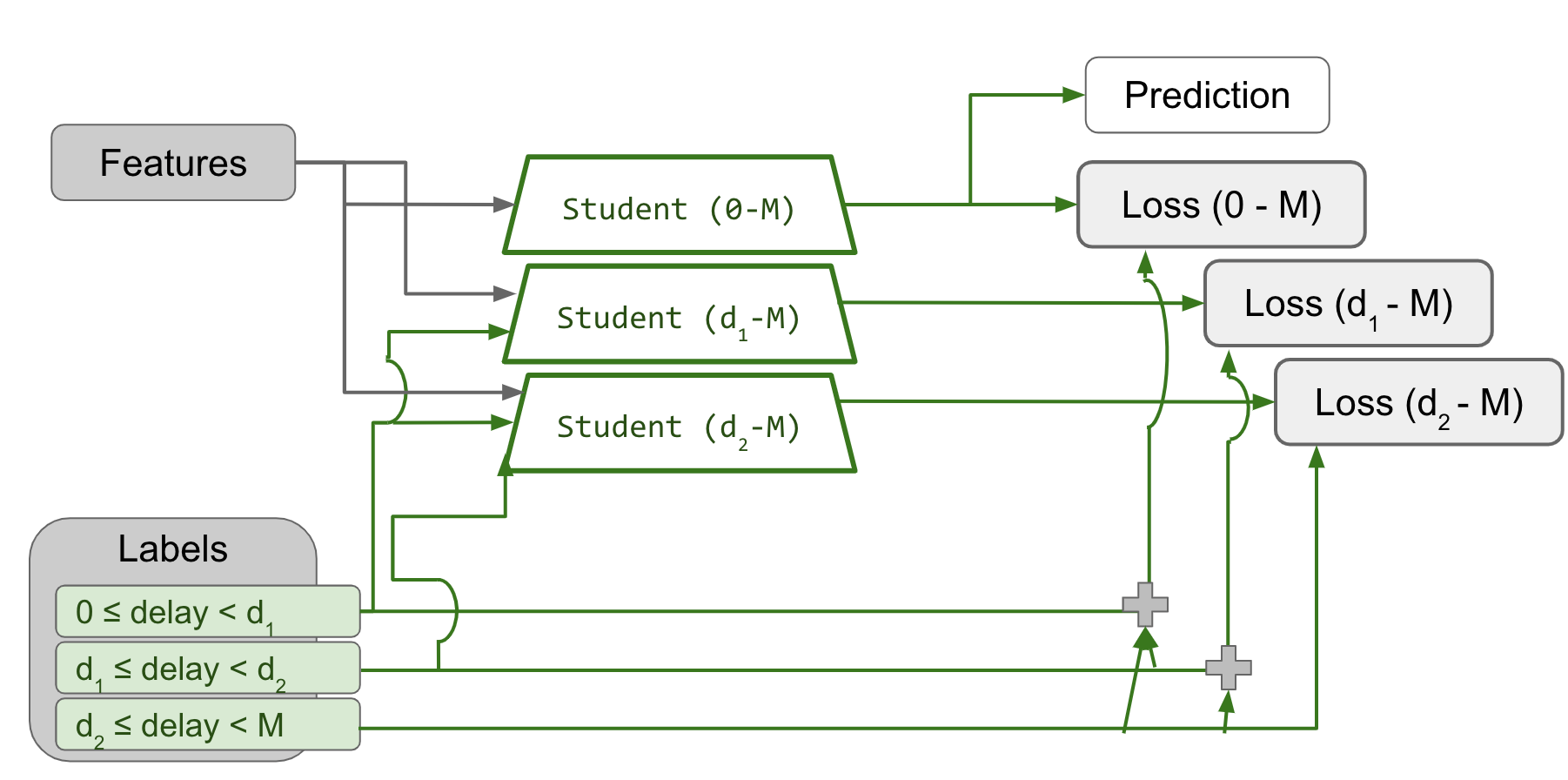}
\caption{Final bare-bones architecture of delay-adjusted prediction model. $M$ specifies the attribution window for the example and $d_i$ are the training delays of each of the auxiliary sub-models. Note that the labels follow thermometer encoding and the labels observed before the training delay of each sub-models are provided as auxiliary information. The labels for examples whose ages are less than their corresponding attribution window are augmented by the sub-model predictions enabling the model to train on incomplete labels.}
\end{figure}

\begin{center}
    \begin{table*}[ht]
    \small
    \begin{tabular}{ | l | l | l | l | l | }
    \hline
    Model & Objective & Label & Examples to train & Features \\ \hline
    $f_n$ & $y_p([t_p+d_n,t_p+M))$ & $y_p([t_p+d_n,t_p+M))$ & $t_p+M\leq t_k$ & $X_p\cup  L_p([t_p,t_p+d_n))$ \\ \hline
    $f_{n-1}$ & $y_p([t_p+d_{n-1},t_p+M))$ & $y_p([t_p+d_{n-1},t_p+d_n))+f_n$ & $t_p+d_{n}\leq t_k$ & $X_p\cup L_p([t_p,t_p+d_{n-1}))$ \\ \hline
    $f_{n-2}$ & $y_p([t_p+d_{n-2},t_p+M))$ & $y_p([t_p+d_{n-2},t_p+d_{n-1}))+f_{n-1}$ & $t_p+d_{n-1}\leq t_k$ & $X_p\cup L_p([t_p,t_p+d_{n-2}))$ \\ \hline
    \ldots & \ldots & \ldots & \ldots & \ldots \\ \hline
    $f_1$ & $y_p([t_p+d_1,t_p+M))$ & $y_p([t_p+d_1,t_p+d_2))+f_2$ & $t_p+d_2\leq t_k$ & $X_p\cup L_p([t_p,t_p+d_1))$ \\ \hline
    $f_0$ & $y_p([t_p,t_p+M))$ & $y_p([t_p,t_p+d_1))+f_1$ & $t_p+d_1\leq t_k$ & $X_p$ \\ \hline
    \end{tabular}
    \caption{Table showing which models are trained, their features/labels and which examples they train on.}\label{table:ModelsFeaturesFilters}
    \end{table*}
\end{center}
\section{Experimental Evaluation}
In this section we will describe the exact setup used in the evaluation.

{\raggedright \textsf{Dataset - } The specific dataset on which we evaluate our solution is app install ads from a commercial mobile app store. Here, all examples are ad clicks and the label is post-click events which happen in the app after the user installs the app. For example, for a ride share app, this could be the total number of rides in a specified time window that the user purchases after installing the app. Like \cite{Chapelle2014A2} we assume last click attribution where any events are attributed to the user's most recent ad click. }

{\raggedright \textsf{Features and models - } We use several categorical features. We embed these categorical features in a dense vector space and take these embeddings and pass them through a fully connected deep neural network. We have several fully connected layers in the neural network. }

{\raggedright \textsf{Optimizer and loss function - } We use the AdaGrad optimizer. We cast the regression problem as a Poisson regression \cite{poissonRegressionBook1} which is one of the standard ways of predicting the expected value of count data, but our results should generalize to other regression formulations. We also tune all the hyperparameters to optimize for maximizing likelihood when trained on mature labels. We use an ensemble of models for each variant to maintain reproducibility of results. \cite{shamir2020antidistillation} }

{\raggedright \textsf{Online training - } We use online training which has been quite popular within many internet companies \cite{McMahan2013} and handles the non-stationarity of data very well. In online training we first start training on the oldest examples and then train on examples in the order of their timestamps until we have completed one pass through the data. We may then continue training on new data as it becomes available as a result of new user interactions. To evaluate model performance during training, we first evaluate performance on each example and then train the model on that example. This gives us an estimate of how the model may have performed if we had chosen to use it in production at any given point in time. }

\subsection{Models compared}
We will compare our new solution against several natural baselines. Along with this we also list several ablations that we conduct on different aspects of the design in our solution.

{\raggedright \textsf{M1: Neglecting delay - } In this model we train on all data and naively use the events or value available after a minimal delay (less than 1 day from click time) as the label. }

{\raggedright \textsf{M2: Train on different delays - } This set of models trains on all data with each model using a different delay from wall time, progressively seeing more of the label and less-recent examples as the delay increases. }

{\raggedright \textsf{M3: Train only on mature data - } Here we throw away most recent examples, for which the label is immature, and only train on examples for which label is completely mature. }

{\raggedright \textsf{M4: Remove thermometer encoding - } We train a model variant where we don't use the proposed thermometer encoding technique. Instead, each sub-model predicts the label in the window $(t_i, t_{i+1}]$ and to obtain the final prediction, we sum the predictions of these sub-models. While this would make the model more expensive to serve, we also show that this makes the accuracy of the predictions worse. }

{\raggedright \textsf{M5: Remove auxiliary information - } In this variant we remove the auxiliary information we provide to the sub-models (the label observed up to the sub-model's prediction period) and use only the features which are available at serving time. }

{\raggedright \textsf{Oracle: Train on complete labels - } In this variant we train a Poisson Regression model on complete labels assuming no delay. This is impossible in practice, since the label will be incomplete until the attribution window is passed. It's used to represent an upper-bound for this prediction task. }

\subsection{Results}
\begin{figure}[h]
\includegraphics[width=8cm]{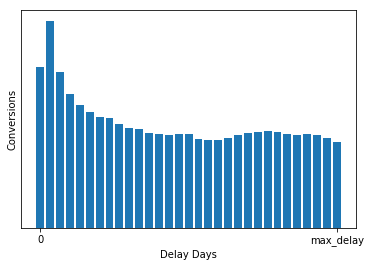}
\caption{Delay Distribution of data}\label{delayDistribution}
\end{figure}

While conversions aggregated over all advertisers may look approximately exponentially distributed, at a per-advertiser level they need not follow any particular class of distributions. Figure \ref{delayDistribution} shows the delay distribution of the data used for evaluation for a specific advertiser, with the Y-axis indicating the number of conversions within the time bucket and the X-axis indicating the progression of time from the observed click.

We evaluate the different variants described above on the data, and compare performance along two dimensions: accuracy and bias. Accuracy is represented by the negative Poisson log-likelihood or the Poisson log loss, since the models considered here are Poisson models predicting conversions per click. The lower the negative Poisson log-likelihood, the more accurate the model variant is at predicting conversions compared to the true final label. The bias defined as the model prediction  divided by the actual true observed label when all the post install conversions have arrived (which we define as ``mature"). Note that the models train on data as it arrives (hence they see only a partially-complete label, depending on the training delay), while during evaluation, we compare the predictions to the true mature label that is attributed to the click at the end of the attribution period. Along with evaluating the models on all data, we also consider slices for new advertising campaigns and from long delay campaigns to show that the proposed model has better performance on partial data compared to the proposed baselines. While we give numbers for improvement in Poisson log loss, we only give qualitative results for bias due to propritary nature of the dataset. We just note that the bias of the new model is $\leq 1\%$ showing that it is completely calibrated. 

\begin{figure}[h]
\begin{center}
\includegraphics[width=4cm]{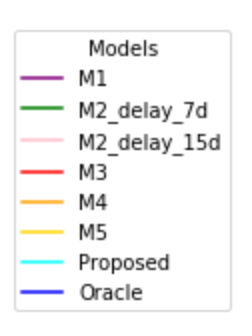}
\end{center}
\caption{Legend for the bias and accuracy plots}
\end{figure}

\begin{figure}[h]
\includegraphics[width=8cm]{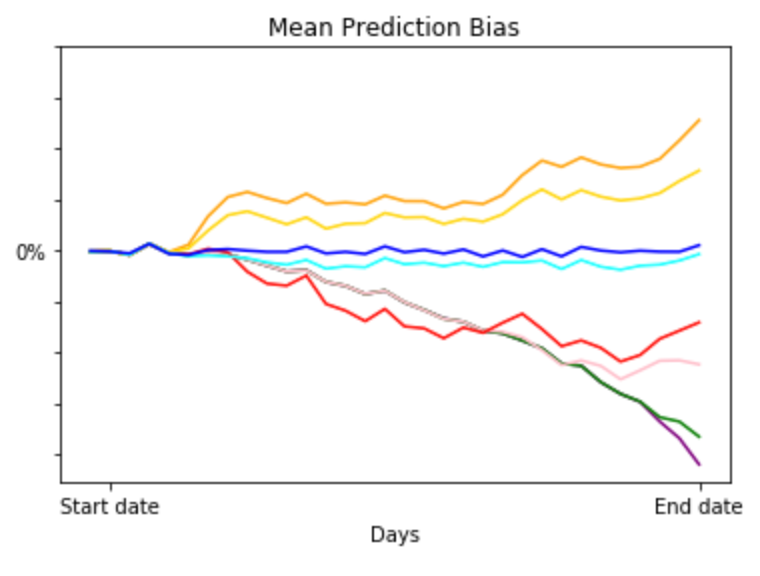}
\caption{Plot of bias of different model variants}
\label{bias:overall}
\end{figure}

\begin{figure}[!htb]
\includegraphics[width=8cm]{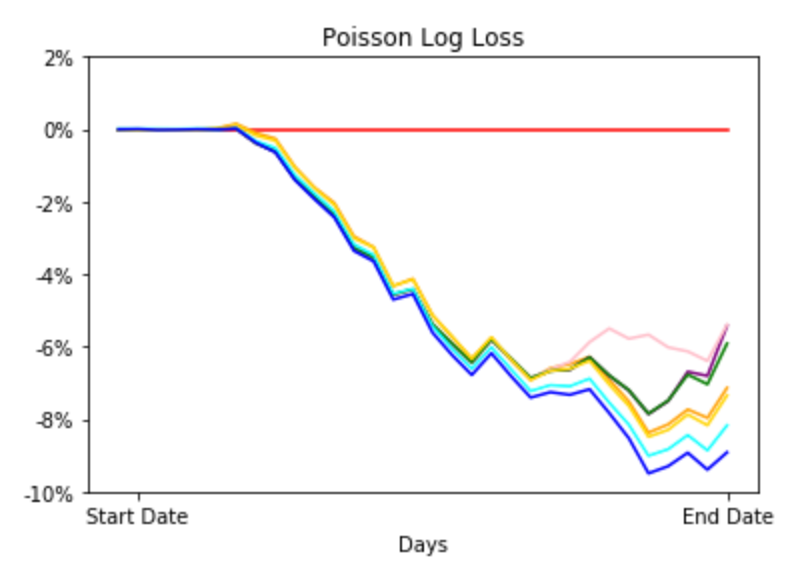}
\caption{Plot showing Poisson Log Loss improvements of different variants with respect to the baseline model M3 training on mature data}
\label{pll:overall}
\end{figure}

Figure \ref{bias:overall} shows prediction bias on all data. As expected, we see that M1 has a significant negative bias since it trains only on a fraction of labels. While all other models have better bias than M1, the proposed model is closest to neutral bias. We can also see that all models are the same when training over mature data, but as each model starts training on examples closer to current time, their performance start diverging due to the sequential nature of training. The models without auxiliary features and thermometer encoding start overestimating labels due to absence of intermediate information and have a positive bias. While the model that trains only on mature data (M3) performs closest to the proposed model on bias, as expected, the accuracy is markedly worse since it trains only on mature data (see Figure \ref{pll:overall}).

Figure \ref{pll:overall} compares Poisson log loss on all data. We can surmise that the more accurate models are expected to have a lower Poisson log loss. The proposed model has the highest improvement on the log loss with respect to the baseline (M3) and therefore has higher accuracy on the test data.  The model variants without auxiliary features (M5) and thermometer encoding (M4) have a higher Poisson Log Loss than the proposed model. The Poisson log loss improvements also show a direct correlation with the training delay, as expected: the smaller the training delay, the greater the improvement in accuracy since the model is training on more recent data. While M4 and M5 have accuracy improvements comparable to the proposed model, they have positive bias which is not seen in the proposed model.

\begin{figure}[!htb]
\includegraphics[width=8cm]{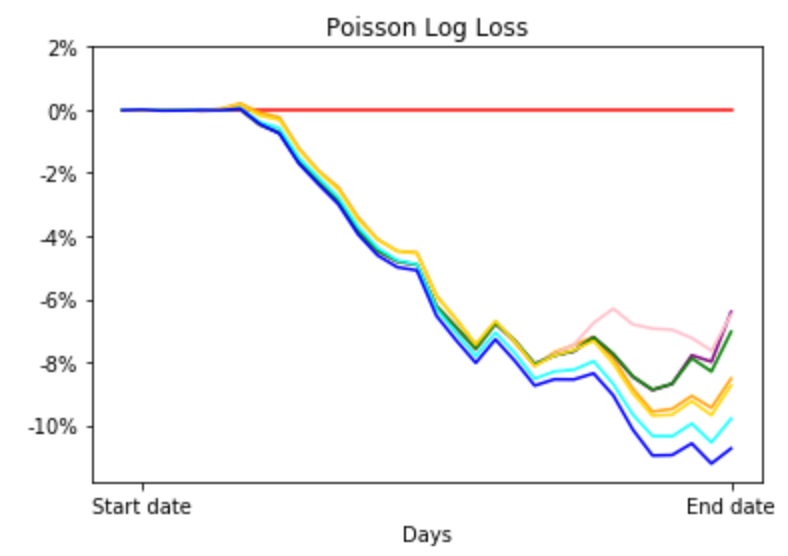}
\caption{Plot of Poisson log loss of different Model variants on high delay (90th percentile) examples compared to the baseline (M3)}
\label{pll:delay}
\end{figure}

\begin{figure}[!htb]
\includegraphics[width=8cm]{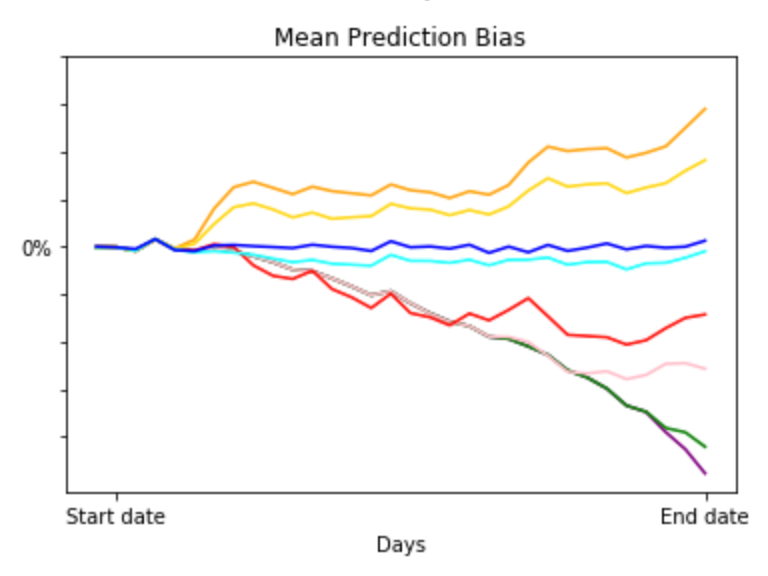}
\caption{Plot of Prediction bias of different Model variants on high delay (90th percentile) examples compared to the baseline (M3)}
\label{bias:delay}
\end{figure}

The difference is more pronounced for high delay campaigns as shown on Figures \ref{pll:delay} and \ref{bias:delay}. This is expected since the proposed model is able to effectively model long tail delay distributions and adjust the label to reflect incomplete data. Other models with shorter delays (M1, M2\textunderscore7d) are much less accurate on these slices and have significantly worse calibration. \\
The other similar variants (M4 and M5) are shown to be less accurate and have positive bias due to lack of intermediate information. 

\begin{figure}[!htb]
\includegraphics[width=8cm]{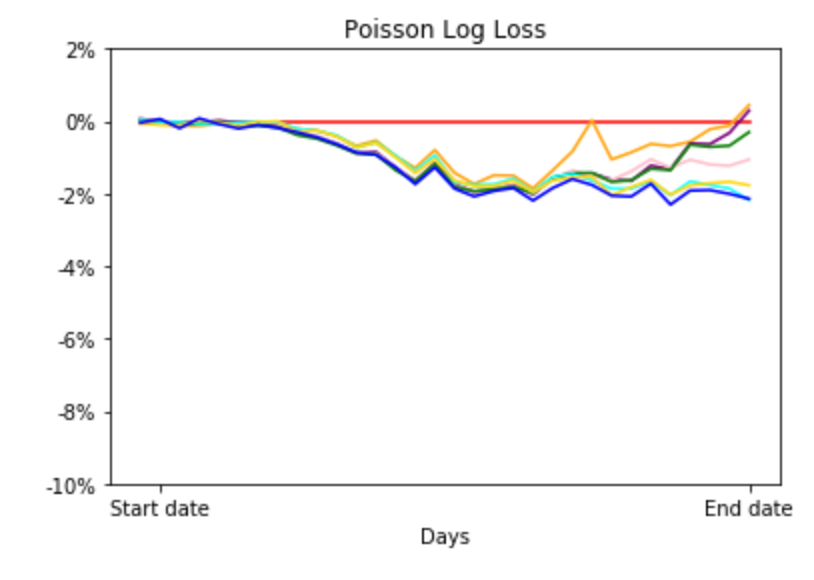}
\caption{Poisson log loss improvements of various model variants on campaigns that are less than 10 days old at start time}
\label{pll:coldstart}
\end{figure}

\begin{figure}[!htb]
\includegraphics[width=8cm]{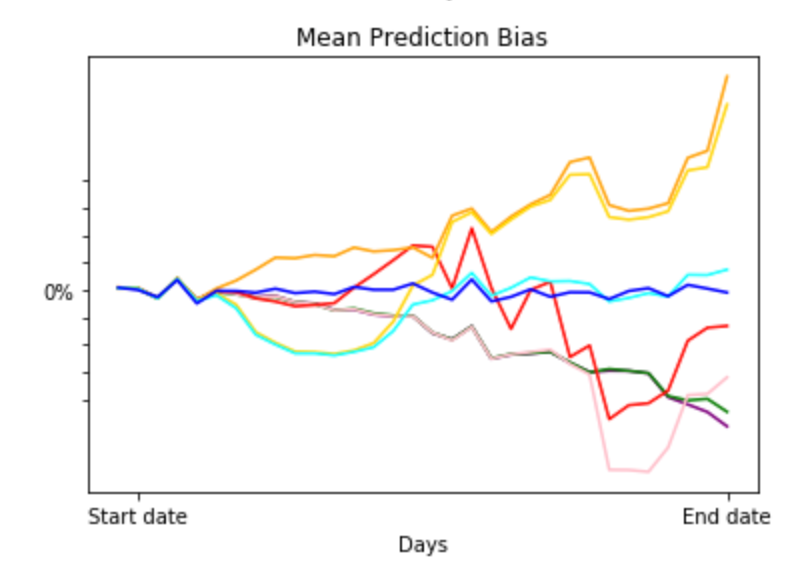}
\caption{Bias of various model variants on new campaigns that are less than 10 days old at start time}
\label{bias:coldstart}
\end{figure}

Another notable improvement of the proposed model is on new campaigns that are only a few days old, where the data is even more limited. By having auxiliary towers and features to predict delay distributions for these campaigns, the proposed model is able to effectively calibrate and adjust to them quickly while maintaining higher accuracy than the other model variants (Figures \ref{pll:coldstart} and \ref{bias:coldstart}). The thermometer encoding is much more important here - as evidenced by the difference in Poisson log loss for new advertisers between M4 and Proposed in Table \ref{table:pll_diff_slices} - since intermediate information and auxiliary features provide crucial signals required to make accurate predictions for new examples. This further bolsters the notion that the proposed model is robust to outliers and limited data compared to the other variants.

\begin{table}[h]
\centering 
\begin{tabular}{|c|c|c|c|}
\hline 
\multicolumn{1}{|p{1cm}|}{\centering Model}
& \multicolumn{1}{|p{1cm}|}{\centering All data}
& \multicolumn{1}{|p{1.5cm}|}{\centering Long delay advertisers}
& \multicolumn{1}{|p{1.5cm}|}{\centering New advertisers}
\\\hline
M3 & 0.0\% & 0.0\% & 0.0\% \\ 
\hline
M1 & -6.6\% & -7.68\% & -0.32\% \\
\hline
M2\_delay\_7d & -6.8\% & -7.97\% & -0.6\% \\
\hline
M2\_delay\_15d & -5.9\% & -7.1\% & -1.13\% \\
\hline
M4 & -7.7\% & -9.13\% & -0.4\% \\
\hline
M5 & -7.92\% & -9.3\% & -1.7\% \\
\hline
\textbf{Proposed} & \textbf{-8.6\%} & \textbf{-10.16\%} &\textbf{-1.81\%} \\
\hline
Oracle & -9.1\% & -10.87\% & -2.0\% \\ [1ex] 
\hline 
\end{tabular}
\caption{Poisson log loss improvements of various model variants on different training slices} 
\label{table:pll_diff_slices} 
\end{table}

\section{Extensions}
In this section we will note how our design is versatile enough to handle several modifications to the problem.

{\raggedright \textsf{Value - } In our experiments we evaluated the setting where we predict expected number of post-click conversions. A simple variant of this problem that is very important in the industry is one in which each conversion can have a different value and the goal is to predict the expected total value of post-click conversions. It is straightforward to see that the whole design works with almost no changes and gives an unbiased estimator in this case.}

{\raggedright \textsf{Handling retractions and restatements - } In all previous solutions to handling conversion delay, the papers assume immutability of conversions which have already appeared. In practice advertisers might want to retract or restate some subset of conversions that they reported. This can be due to customers returning an item or a conversion having been found to be fraudulent. It is relatively straightforward to modify our solution to handle retractions and restatements while still obtaining an unbiased estimator. To do this, we will need to split a conversion across different time buckets: a +1 in the bucket in which it happened and -1 in the bucket in which it was retracted. This will define consistent random variables, but can make the label negative in some of the time buckets which cannot be handled by Poisson regression. A simple fix to handling negative labels is to split the label in any time bucket into a positive portion and a negative portion, and to have separate outputs from the neural network to predict each of these.}
\section{Conclusion}
In this paper we introduced a way of handling delayed feedback in conversion optimizer models with many conversions per click. We showed experimentally that it does better than several other solutions as well as via ablation showed that all ideas introduced are necessary. We also showed that it is robust to outliers and limited data. Our solution is likely to be useful for problems in other domains with delayed feedback. 
\section{Acknowledgments}
This authors would like to thank Samuel Ieong, Eu-Jin Goh, and Camille Wormser for their support and valuable inputs. 

\bibliography{main}
\bibliographystyle{custom2019}

\end{document}